# MOELA: A Multi-Objective Evolutionary/Learning Design Space Exploration Framework for 3D Heterogeneous Manycore Platforms

Sirui Qi
Dept. of Electrical and Computer Engg.
Colorado State University
Fort Collins, USA
alex.qi@colostate.edu

Yingheng Li
Dept. of Computer Science
University of Pittsburgh
Pittsburgh, USA
yil392@pitt.edu

Sudeep Pasricha, Ryan Gary Kim
Dept. of Electrical and Computer Engg.
Colorado State University
Fort Collins, USA
{sudeep, ryan.g.kim}@colostate.edu

*Abstract*— To enable emerging applications such as deep machine learning and graph processing, 3D network-on-chip (NoC) enabled heterogeneous manycore platforms that can integrate many processing elements (PEs) are needed. However, designing such complex systems with multiple objectives can be challenging due to the huge associated design space and long evaluation times. To optimize such systems, we propose a new multi-objective design space exploration framework called MOELA that combines the benefits of evolutionary-based search with a learning-based local search to quickly determine PE and communication link placement to optimize multiple objectives (e.g., latency, throughput, and energy) in 3D NoC enabled heterogeneous manycore systems. Compared to state-of-the-art approaches, MOELA increases the speed of finding solutions by up to 128×, leads to a better Pareto Hypervolume (PHV) by up to 12.14× and improves energy-delay-product (EDP) by up to 7.7% in a 5-objective scenario.

## I. INTRODUCTION

The growing use of deep neural networks, graph analytics, and Big Data computing applications generates an increasing demand for computational power and energy-efficient hardware platforms. To meet these demands, three-dimensional (3D) heterogeneous manycore systems that integrate multiple processing elements (PEs) of different types (e.g., CPUs and GPUs) on each layer in the 3D stack can provide high performance while being energy efficient.

To facilitate communication in such heterogeneous manycore systems, a 3D network-on-chip (NoC) is typically implemented to transfer data between cores and main memory [1]. However, designing the topology and configuration of these 3D NoCs for heterogeneous manycore systems under multiple design constraints is very challenging. For example, CPUs require the NoC to provide low memory latency accesses while GPUs require high data throughput. These requirements can often conflict with one another, creating congestion in the network and reducing the performance of the overall system. The need to balance thermals and improve energy efficiency further increases the exploration complexity. Thus, the design of 3D heterogeneous manycore systems is a multi-objective optimization (MOO) problem.

Evolutionary algorithms (EAs) [2], [3] such as NSGA-II [4] and MOEA/D [5] have been proposed to solve MOO problems in prior work. In EAs, evolution-inspired operations are applied to a group (population) of solutions. EAs typically perform well to find good solutions along the Pareto front [6] but usually take a long time to converge and do not scale well as more objectives are added and design space sizes increase. To improve design space exploration times, machine learning (ML)-guided local search approaches such as MOOS [7] and MOO-STAGE [8] have been presented. By using past search histories, these approaches learn and guide future searches towards promising areas of the design space. However, these approaches rely on greedy local search techniques that do a good job at speeding up the design space search, but at the cost of losing diversity along the Pareto front.

In order to achieve fast exploration speed, high quality of solutions, and diversity along the Pareto front, in this paper, we propose a novel hybrid exploration approach that combines the benefits of both EAs and ML-guided techniques. The novel contributions in this work include:

- We propose a novel hybrid multi-objective evolutionary /learning algorithm called MOELA to solve large-scale MOO problems, such as the 3D NoC based heterogeneous manycore system design problem.
- We compare and contrast MOELA against the state-of-the-art algorithms, such as MOEA/D [5] and MOOS [7], in terms of speedup and solution quality.
- We demonstrate that MOELA improves the speed up to 128× while achieving a better Pareto Hypervolume (PHV) [6] by up to 12.14×, and improves energy-delay-product (EDP) by up to 7.7% in a 5-objective scenario on average over state-of-the-art methods for a MOO problem with five objectives.

By using both EAs and ML-guided techniques, MOELA can rely on the ML-guided algorithm to find a few high-quality solutions to supplement the population while relying on the EAs to maintain a diverse population. Over multiple iterations, we find that MOELA's hybrid approach finds good solutions in a shorter time and discovers better solutions than either EAs or ML-guided approaches alone.

The rest of the paper is organized as follows. Section II discusses relevant prior work. In Section III, we formulate our design problem. Section IV introduces the components and structure of the MOELA framework. Sections V and VI discuss the experiment results and present our conclusions.

## II. RELATED WORK

### A. MOO with Evolutionary Algorithms

EAs have been proposed to solve MOO problems using evolution-inspired operations, such as crossover, mutation, and survival, on a population of solutions. These operations allow EAs to explore across a multi-objective design space and allow parameters that lead to the best solutions to survive across generations of populations. NSGA-II [4] has been a popular EA in computer system design problems. NSGA-II uses non-dominated sorting and crowding distance to select the next population that is closest to the Pareto front when implemented to perform design space exploration for application mapping in a heterogenous manycore system.

MOEA/D [5] is another popular EA that decomposes the problem into multiple subproblems to cover multiple directions in the design space. Compared with NSGA-II, MOEA/D is generally faster while acquiring similar results. In [9], MOEA/D was used to map Intellectual Property (IP) cores to balance traffic load and energy consumption in the NoC. In [10], NSGA-II is used for resource allocation in a high-performance computing data center. Also inspired by NSGA-II, several MOO design problems have been addressed for multiprocessor system-on-chips (MPSoCs). These MOO problems consider challenges with process variations, energy consumption, reliability, and performance [11]-[13]. Furthermore, NSGA-II has inspired researchers to solve problems involving trade-offs between performance and power dissipation in NoCs [14], [15]. It is important to note that all of these design problems are much less complex than the problem we will discuss in-depth in Section III.

*B. MOO with ML-Guided Algorithms*

Local search techniques attempt to find good solutions by changing the current solution(s) by a small amount and accepting new solutions by following some heuristic. Although local searches can find good solutions over time, they do not use any information from past searches when performing repeated searches. ML-guided local searches attempt to remedy this and speed up search times. In [8], inspired by the STAGE [16] approach, MOO-STAGE was proposed to learn from past search history and find a good starting point for MOO local searches. MOO-STAGE was shown to perform over 10× faster and find solutions with 9.6% better energy-delay product (EDP) over traditional MOO local search approaches for a four-objective 3D NoC heterogeneous manycore design problem. Another framework, MOOS [7], attempts to use learned information to adjust the local search direction. The authors of the work found that MOOS was able to improve the search time by 3.4× over MOO-STAGE while finding 13.9% better quality solutions for the same four-objective 3D NoC heterogeneous manycore design problem. Although these results are impressive, the results still take days to generate. Additionally, in our experience, the solution quality obtained from the framework also deteriorates as we scale up system size and the number of objectives. In this paper, we address these shortcomings as part of our proposed novel MOELA framework for MOO problems.

### III. DESIGN PROBLEM FORMULATION

For a 3D NoC-based heterogeneous manycore system design problem, we are provided with a $N \times N \times Y$ tile system with $Y$ layers of $N \times N$ tiles. Each of these tiles accommodate one PE (e.g., CPUs, GPUs, or last level caches (LLCs) with a memory controller). To allow the tiles to communicate with one another, each tile has a NoC router that can be interconnected with other tiles via a communication link. This communication link could be planar links that connect same-layer routers or through-silicon via (TSV) links that connect tiles on neighboring layers in the 3D stack. We are provided with $L$ links (planar and vertical) for the entire system. An example $3 \times 3 \times 3$ system is shown in Fig. 1. From application profiling, we also obtain the communication frequency between tile $i$ and tile $j$ ($f_{ij}$) and the average power consumption for each PE. Then, the design problem is to determine the location of the $L$ planar links and the ($N \times N \times Y$) tiles that optimize the objectives subject to the constraints, both of which are described next.

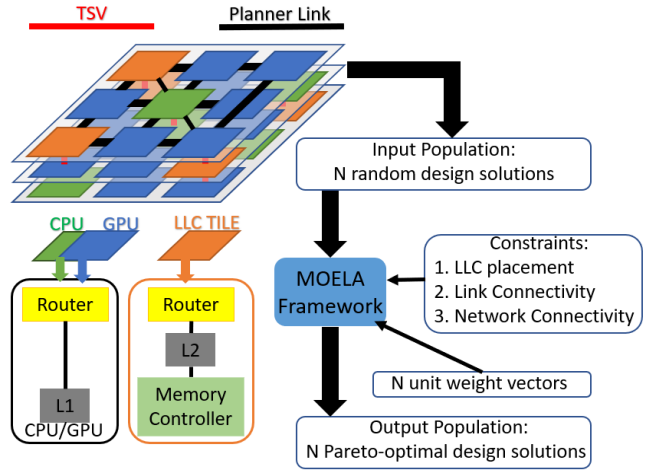

Fig. 1. Example of 3-layer 27-tile 3D NoC System.

To maintain a practically feasible system, we define the following set of constraints:

- To ensure that each router is able to communicate with any other router in the network, we ensure that every tile has a network path (connectivity) to every tile in the system.
- The total number of links (*L*) is fixed. Links can be placed anywhere between tiles if connectivity is guaranteed.
- We restrict the length of each planar link to 5 units (1 unit = distance between adjacent tiles in the same layer) and the maximum number of links connected to any single router to 7 to reduce NoC router area and wiring overheads.
- The maximum number of vertical links between adjacent tiles in adjacent layers is set to be 1.
- Due to interfacing constraints, since the tiles with LLCs contain memory controllers that must be able to access the main memory, the LLC tiles must be placed along the edge of the dies in the 3D stack.

We define the following five objectives to consider performance, energy, and thermals as part of our MOO:

**1) Mean of traffic:** To minimize the average traffic across the network, we want to reduce the mean link utilization:

$$Mean = \frac{1}{L}\sum_{k=1}^{L} u_k = \frac{1}{L}\sum_{k=1}^{L}\left(\sum_{i=1}^{A}\sum_{j=1}^{A} f_{ij} \cdot p_{ijk}\right) \quad (1)$$

Here, $u_k$ is the utilization of the $k^{th}$ link of $L$ total links, $A$ is the number of tiles, $f_{ij}$ is the frequency of communication between tiles $i$ and $j$, and $p_{ijk}$ is an indicator function for if communication between tiles $i$ and $j$ uses link $k$.

**2) Variance of traffic:** In tandem with the first objective, we also focus on reducing the traffic hotspots, i.e., links with higher utilizations, by reducing the variance of link utilization (*Variance*) to improve network throughput for GPU traffic:

$$Variance = \frac{1}{L}\sum_{i=1}^{L}(u_i - Mean)^2 \quad (2)$$

**3) CPU latency:** CPUs are especially sensitive to memory access latency. Therefore, we include an objective that focuses on CPU-LLC latency. For $C$ CPUs and $M$ LLCs, we model the average CPU-LLC latency using the following equation:

$$Latency = \frac{1}{CM}\sum_{i=1}^{C}\sum_{j=1}^{M}(rh_{ij} + d_{ij}) \cdot f_{ij} \quad (3)$$

In the above equation, $r$ is the number of router pipeline stages in each router, $h_{ij}$ is the number of network hops between CPU $i$ and LLC $j$, and $d_{ij}$ is the total link delay.

**4) Energy consumption:** To reduce the overall energy consumption of a heterogeneous NoC, we need to minimize the sum of link and router energy:

$$E = \sum_{i=1}^{A}\sum_{j=1}^{A} f_{ij}(\sum_{k=1}^{L} p_{ijk} d_k E_{link} + \sum_{k=1}^{R} r_{ijk} E_r P_k) \quad (4)$$

$E_r$ and $E_{link}$ denote the average router logic energy per port and the average link energy per flit, respectively, $d_k$ represents the physical length of link $k$, and $P_k$ represents the number of ports at router $k$. $p_{ijk}$ and $r_{ijk}$ are defined to indicate whether a link or router $k$ is utilized to communicate between tile $i$ and tile $j$, respectively. $R$ is the total of routers.

**5) Thermal:** To accurately estimate the peak temperature of a core, we use the fast approximation model presented in [17]. Our system can be divided into $N \times N$ single-tile stacks, with $Y$ layers. The temperature of a core within a single-tile stack $n$ located at layer $k$ from the sink ($T_{n,k}$) is given by:

$$T_{n,k} = \sum_{i=1}^{k}(P_{n,i}\sum_{j=1}^{i} R_j) + R_b \sum_{i=1}^{k} P_{n,i} \quad (5)$$

Here, $P_{n,i}$ is the average power consumption of the core $i$ layers away from the sink in single-tile stack $n$, $R_j$ is the vertical thermal resistance, and $R_b$ is the thermal resistance of the base layer on which the dies are placed. The horizontal heat flow can be also estimated by the maximum temperature difference in the same layer k ($\Delta T(k)$) according to [17]:

$$\Delta T(k) = \max_n T_{n,k} - \min_n T_{n,k} \quad (6)$$

Hence, we can acquire the overall thermal model by combining the heat models represented by (5) and (6):

$$T = \max_{n,k} T_{n,k} \times \max_k \Delta T(k) \quad (7)$$

## IV. MOELA FRAMEWORK

In this section, we discuss our proposed novel MOO algorithm called MOELA which can efficiently solve large and complex MOO problems. Fig. 1 gives a high-level overview of the MOELA framework to explore the design space for a 3D NoC-based heterogeneous manycore system. The input to the MOELA framework is $N$ generated solutions that have a randomly selected tile and link placement. This framework is also provided with constraints and $N$ weight vectors that have the same number of dimensions as the objective space, and with the goal of ensuring that the design space exploration is performed in evenly dispersed directions. The output of the MOELA framework is $N$ Pareto-optimal designs, which enable useful trade-offs among the multiple design objectives and preserve diversity in the solution space.

### A. MOELA: Overview

There are three major areas that can be improved for prior design space exploration algorithms: 1) convergence time, 2) anytime solution quality, i.e., the best solution at any point in time, and 3) coverage of the pareto front (solution diversity). Our proposed MOELA framework helps to decrease the convergence time while improving the design quality of complex MOO problems such as the 3D heterogeneous NoC problem described in Section III. MOELA works by iterating over two integrated steps: (1) an EA that tries to advance the Pareto front while maintaining diversity and (2) a local search method for focused search that greatly advances the Pareto front in a particular direction in the objective space.

In prior local search work, such as MOO-STAGE [8], the update principle of the local search aims to maximize the Pareto Hypervolume (PHV). For a set of solutions, the PHV is a metric that measures the hypervolume of the solution space that this set of solutions dominates, i.e., equal or better in all objectives and better in at least one objective. The evaluation function in MOO-STAGE is also trained to predict the PHV of the local search trajectory. This helps the search provide a set of solutions that span the Pareto front. However, the repeated calculations of PHV during local search can lead to large computational overhead especially in high dimension search spaces. In addition, this also results in a complex learned evaluation function that needs to consider the current population of solutions in addition to the search trajectory. Therefore, we propose a novel machine learning (ML) guided local search method that uses a decomposition method in both its local search and learned evaluation function. This new ML-guided local search in MOELA improves the effectiveness of single node local search, allowing the search to focus on the best solution in a limited search space in the direction of the unit weight vector (the same weight vector is used in the weighted sum). MOELA further aims to preserve diversity of solutions with EAs and their genetic operators to increase PHV in the high dimensional space.

In summary, MOELA utilizes its ML-guided local search to reach optima in multiple directions and then implements local-search-friendly EA to avoid local optima and preserve solution diversity. Fig. 2 shows an overview of MOELA while Algorithm 1 describes the main steps within MOELA.

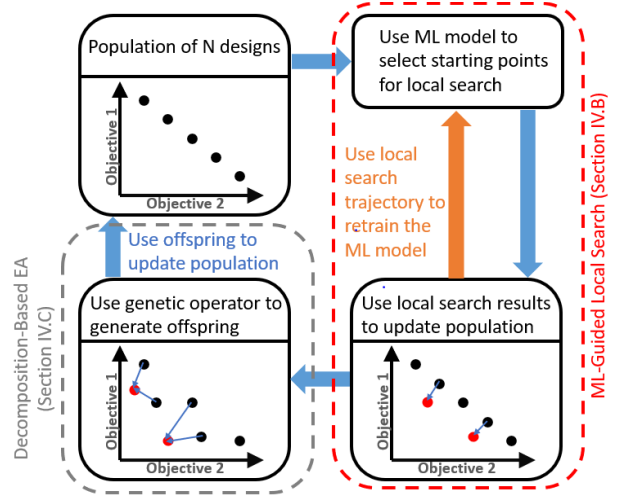

Fig. 2. Overview of our proposed MOELA framework.

In MOELA, first, as done in the decomposition-based EAs, we decompose the MOO problem into $N$ single-objective optimization sub-problems where each sub-problem is defined by a particular weight vector. Then, we create the initial population $P$ with $N$ different designs and associate each design with one of the sub-problems, and hence, one of the weight vectors. These weight vectors are evenly spread out and point at different directions across the objective space. For example, with $N = 11$ and the number of objectives = 2, the weight vector set $W = \{[0,1], [0.1,0.9], ..., [1,0]\}$. Then, we perform a series of local searches on this population. Based on our investigations, we start with the local search since we find that using a local search before EAs provides the best results.

Except early iterations, in each iteration of MOELA, we first use an ML-guided local search to find the local optima for some of the subproblems (Algorithm 1, lines 4-9). More details of the ML-guided local search are given in Section IV.B and Algorithm 2. This allows MOELA to target the local searches at specific areas of the objective space independently, which guarantees a more thorough search. After the local searches, the EA can then use the results of the local searches to improve the rest of the population using genetic operations and mutations. MOELA will iterate over both local search and EA stages to find better solutions than either of these alone. In this process, local searches increase the convergence speed while EA maintains the diversity of the population.

---

**ALGORITHM 1**: MOELA

**Input:** $gen, N, iter_{early}, n_{local}, \delta$
**Output:** Population $P$ (Final N designs)

1: **Initialization:**
   **Weight Vector Set** $W = \{w_1, ..., w_N\}$, all weight vectors are evenly spread.
   **Population** $P = N$ random designs $\{p_1, ..., p_N\}$, with random weight vector $w \in W$.
   **Neighbor Set** $\mathbb{N}_i = \{p_{i,1}, \cdots, p_{i,T}\}$, $T$ designs in $P$ with the closest assigned weight vectors to $p_i$
   **Training Set** $S_{train} \leftarrow \emptyset$
   **Reference Point** $z = [o_1, ..., o_M]$ where $o_i$ is the objective value of the $i$th objective out of $M$ objectives.
2: **for** $i = 0$ to $gen$ **do**
3:     **if** $i < iter_{early}$: $P_{start} \leftarrow random(n_{local}, P)$
4:     **else**: $P_{start} \leftarrow MLguide(Eval, n_{local}, P)$
5:     **for** $p, w$ in $P_{start}$ **do**
6:         $(p_{new}, S_{traj}) \leftarrow localSearch(p, w, z)$
7:         $S_{train} \leftarrow S_{train} \cup S_{traj}$
8:         $P \leftarrow updatePopulation(P, p_{new}, W)$
9:     **end for**
11:     Train evaluation function:
           $Eval \leftarrow MLtrain(S_{train})$
12:     $P \leftarrow EA(P, \mathbb{N}, \delta, Z, W)$
13: **end for**
14: **return** $P$

---

In the following subsections, we discuss the details of the decomposition-based EA and the ML-guided local search steps used in our approach, and the amendments needed for these to work in the MOELA framework.

### B. MOELA: ML-Guided Local Search

Inspired by the success of ML-guided MOO algorithms using local search for 3D heterogeneous NoC design [2], [8], we use an ML-guided local search to boost some of the populations before executing EAs. Although the traditional optimization algorithms are widely used to solve a variety of problems, they still have some shortcomings, namely, slow convergence speed to the Pareto front. A local search module not only improves the overall population quality, but also provides much better individuals for EA to choose as parents and then generate better offsprings. These outstanding individuals, when generated by a local search module, will be much better than other individuals in the same generation. From our experience, we observe that even a few outstanding individuals in a huge population can lead to a much better next generation population when utilized by the same EA.

In MOELA, instead of randomly selecting starting points for the local search as in standard local searches, we use a modified version of STAGE [16] as a guide to select the most promising designs from the current Pareto front for the upcoming local search. STAGE does this by training an ML model using past local search history to predict the potential of different designs as starting points to the search for single objective problems. In MOELA, to reduce the MOO to a single-objective problem and better interface with the decomposition-based EA (Section IV.C), we propose a new guide approach (Algorithm 2) that uses both the decomposition method and machine learning in upcoming local search starting point selections.

For each iteration, we choose $n_{local}$ designs in the population as starting points for our local search. Since there is not enough data for training in the early iterations, we randomly select the $n_{local}$ designs for the local searches in the first $iter_{early}$ iterations (Algorithm 1, line 3). MOELA then performs independent local searches from these $n_{local}$ designs. Unlike prior ML-guided local searches and to maintain consistency with the decomposition approach used in the EA, the local search includes the reference point $z$ and the design's weight vector. This directs the local search towards the reference point $z$, with emphasis on certain objectives depending on the weights. The local search minimization function is:

$$minimize \ g(Obj|w, z) = \Sigma_{i=1}^{M}\{w_i|Obj_i - z_i|\} \quad (8)$$

where $Obj_i$ is the $i^{th}$ objective value out of the $M$ objectives of a design. We use a simple greedy descent approach for our local search. The local search returns the search trajectory (each design visited) and the final value of (8) and they are recorded in the training set $S_{train}$ (aggregated over iterations). The population is updated if a better solution is found for that weight. MOELA then attempts to learn an evaluation function $Eval$ that maps each design's parameters and weight to the result of the search (Eq. (8)) using the aggregated training set $S_{train}$ (Algorithm 1, line 11). To create $Eval$, we employ a random forest model, which is an ensemble model that uses the average output from a collection of decision trees to help reduce overfitting. Random forests have previously performed well for MOO-STAGE [8], however, any sufficiently expressive model would work here.

---

**ALGORITHM 2**: MLguide

**Input:** $Eval, n_{local}, P$
**Output:** $P_{local}$

1:   **for** $p_i$ to $P$ **do**
2:     $e_i = Eval(p_i)$
3:   **end for**
4:   $E = \{e_1, ..., e_N\}$
5:   **return** $n_{local}$ designs in $P$ and their weights with lowest $e_i$

---

After running $iter_{early}$ iterations, MOELA starts to evaluate all $N$ designs in the population using $Eval$ to find the most promising $n_{local}$ designs to perform the local search (Algorithm 2). Essentially, the algorithm attempts to learn a regressor that can predict how much a design can improve towards the reference point in a local search. Note that the local search approach in MOELA is different than MOO-STAGE and MOOS which attempt local searches on the entire archive of solutions for all objectives. By proposing a new ML-guided decomposition-based local search, we avoid costly PHV calculations, learn an $Eval$ function that is unrelated to the population of solutions, and better interface with the following decomposition-based EA step.

### C. MOELA: Decomposition-Based EA

In a decomposition-based EA, the single-objective optimization sub-problems are defined with a set of $N$ uniformly spread weight vectors $W = \{w_1, ..., w_N\}$ by the Tchebycheff approach [18]:

$$minimize \ g(x|w, z) = \max_{1 \le i \le M}\{w_i|Obj_i(x) - z_i|\} \quad (9)$$

where $g$ is the scalar optimization problem, $M$ is the number of objectives, $Obj_i(x)$ is the $i^{th}$ objective value of input $x$,

and $Z = \{z_1, ..., z_m\}$ is the reference point defined as the minimum value of all the objectives for the objective space of population $P$. Given a weight vector $w_i$, a lower Tchebycheff value $g(x|w,z)$ means a better design is found for the $i^{th}$ subproblem. In EA, an offspring is generated from two parent designs using a genetic operator (GO). This GO attempts to take two parent designs and aims to create offsprings that contain the best attributes of both parents. If the parents are attempting to optimize similar sub-problems (similar weights) the offspring has a higher chance to improve the sub-problem than parents with dissimilar sub-problems. This implies that two designs with close weight vectors in MOELA will make better parents. Hence, we frequently choose parents for the EA from a neighborhood, which is defined as the $T$ closest sub-problems based on the Euclidean distance of the weight vectors. MOELA use the same neighborhood setting as [5] to improve the quality of the offspring from EA. Here, we set a probability $\delta$ that chooses the sub-problem's neighborhood as the parent pool $Q$ and $1 - \delta$ probability that choose from all $1, \cdots, N$ designs to improve diversity of the population. After generating a new design $P_{new}$, MOELA updates the parent pool $Q$ with the consideration of assigned weights $W$:

$$P \leftarrow \text{Update Population}(P, P_{new}, W) \quad (10)$$

A design of the population front will be updated if the Tchebycheff value of $P_{new}$ is lower than that of the other designs in $Q$. Due to the weight vectors and neighborhood, MOELA tries to generate an offspring targeting a specific area of the objective space, which is different from other EA that randomly choose parents to generate offspring.

## V. EXPERIMENTAL RESULTS

### A. Experimental Setup

In our experimental studies, we consider a 4×4×4 tile system that consists of 40 NVIDIA Maxwell GPU cores, 8 x86 CPU cores, and 16 LLCs. The CPUs operate at 2.5 GHz while the GPUs operate at 0.7 GHz. The memory system uses a MESI two level cache coherence protocol. Each CPU and GPU have a private L1 data and instruction cache of 32 KB each. Each LLC consists of 256 KB memory. To allow communication between these tiles, we have allocated 96 planar communication links (the same number as an equivalent 3D mesh) and 48 TSVs that can be used to connect tiles. To obtain the traffic patterns ($f_{ij}$) and average power profiles needed for MOELA, we use Gem5-GPU [19] and GPGPU-Sim [20], McPAT [21], and GPUWattch [22]. We also use these simulators to generate the final EDP results from the solutions found by MOELA. To generate the thermal profiles, we use 3D-ICE [23]. The values of $R_j$ and $R_b$ are also obtained using 3D-ICE.

We use benchmarks from the Rodinia heterogeneous (CPU+GPU) computing suite [24]. Rodinia is by far the most available CPU-GPU heterogenous computing benchmark, which has applications in Machine Learning, Bioinformatics, Data Mining, Pattern Recognition etc. Here, we pick seven different applications from Rodinia that represent a variety of computing scenarios to test our design framework: Back Propagation (BP), Breadth-First Search (BFS), Gaussian Elimination (GAU), Hotspot3D (HOT), PathFinder (PF), Streamcluster (SC) and SRAD. We evaluate MOEA/D [5], MOOS [7], and MOELA across these benchmarks in terms of search time, PHV, EDP, and peak temperature. The design space exploration is performed on a server with an AMD CPU (EPYC 7763 @ 2.45 GHz) with 32GB of RAM. The hardware simulations are performed on a server with an AMD Ryzen 7 3700X @ 3.6GHz machine with 32 GB of RAM. The code for MOELA has been made available on Github [25].

### B. MOELA Parameters

In the experiments, we use the following parameters for MOELA to obtain results:
- $N = 50$, the size of the population
- $iter_{early} = 2$ (Algorithm 1)
- $gen = 1000$ was found to be sufficient for all algorithms to converge (Algorithm 1)
- $\delta = 0.9$ (Algorithm 1)
- $|S_{train}| \leq 10K$, we limit the training set to the most recent 10K samples. We find that this selected value does not adversely affect $Eval$'s accuracy.
- $T_{stop}$ = 48 hour, this is the maximum running time bound for all three algorithms that are compared.

### C. Metrics

We use three metrics to compare the performance and speed of different algorithms similar to the metrics in [7]:

1) **Speed-up factor:** The speed-up factor is defined as $T_{convergence}/T_{MOELA}$. Here, $T_{convergence}$ is the time when each algorithm reaches its convergence performance (the improvement of PHV is smaller than 0.5% in 5 iterations) $T_{MOELA}$ is the time when MOELA achieves the same quality of design as the other two algorithms (MOEA/D and MOOS).
2) **PHV improvement:** PHV improvement from MOEA/D and MOOS to MOELA at the maximum stop time $T_{stop}$.
3) **EDP improvement:** EDP improvement from MOEA/D and MOOS to MOELA at the maximum stop time $T_{stop}$.

TABLE I. SPEED-UP OF MOELA COMPARED TO MOEA/D AND MOOS

| App | MOEA/D | | | MOOS | | |
|---|---|---|---|---|---|---|
| | *3-obj* | *4-obj* | *5-obj* | *3-obj* | *4-obj* | *5-obj* |
| BFS | 1.31 | 59.68 | 3.11 | 24.58 | 1.31 | 128.18 |
| BP | 10.88 | 24.72 | 1.40 | 35.29 | 10.89 | 10.64 |
| GAU | 15.51 | 78.8 | 5.07 | 39.42 | 15.52 | 58.73 |
| HOT | 68.41 | 20.94 | 24.85 | 13.65 | 68.41 | 43.90 |
| PF | 32.30 | 20.43 | 7.54 | 36.53 | 32.30 | 7.32 |
| SRAD | 31.92 | 25.83 | 9.28 | 27.19 | 31.92 | 15.81 |
| Average | 34.59 | 121.24 | 8.91 | 38.67 | 34.55 | 38.83 |

TABLE II. PHV GAIN OF MOELA COMPARED TO MOEA/D AND MOOS

| Application | MOEA/D | | | MOOS | | |
|---|---|---|---|---|---|---|
| | *3-obj* | *4-obj* | *5-obj* | *3-obj* | *4-obj* | *5-obj* |
| BFS | 18% | 2% | 326% | 20% | 129% | 18% |
| BP | 32% | 9% | 15% | 8% | 30% | 30% |
| GAU | 22% | 41% | 101% | 25% | 128% | 19% |
| HOT | 8% | 27% | 38% | 0.2% | 10% | 24% |
| PF | 18% | 96% | 19% | 3% | 66% | 20% |
| SRAD | 24% | 36% | 124% | 3% | 1114% | 12% |
| Average | 20% | 35% | 104% | 18% | 247% | 21% |

### D. Results

To observe each algorithm's performance across a variety of objectives, we present the results for three different scenarios: 3-obj (objectives 1-3 in Section III), 4-obj (objectives 1-4), and 5-obj (objectives 1-5). The algorithms' results are shown in Table I, which shows speed up factor for MOELA vs. MOEA/D and MOOS, and Table II, which shows % PHV improvement for MOELA vs. MOEA/D and MOOS. From the tables, we can observe the effectiveness of MOELA which outperforms both competitors. MOEA/D needs significantly larger running time to reach the same PHV as MOOS or MOELA. MOELA's PHV improves much quicker compared to MOEA/D's. MOELA is 8.91x faster and leads to 104% better PHV on average than MOEA/D for the 5-obj case. For BFS, MOELA achieves a very high PHV gain of 326% better than MOEA/D in the 5-obj scenario.

Compared to MOOS, MOELA is 38.83x faster and leads to 21% better PHV on average for the 5-obj case. The improvement comes from the EA and ML models used within our proposed MOELA framework. The EA part of MOELA maintains diversity in the population, which gives MOELA greater ability to jump out of potential local optima compared with MOOS. Meanwhile, the ML-guided local search can ensure that MOELA always performs productive local search and avoids time overhead on sub-optimal local searches as MOOS does. On the other hand, the ML-guided local searches provide outstanding individuals for MOELA to perform EA.

In summary, MOELA is faster than fast MOO approach like MOOS with even better solution quality. Meanwhile, MOELA also provides better quality solutions than EA-based approaches like MOE/AD while taking much less search time.

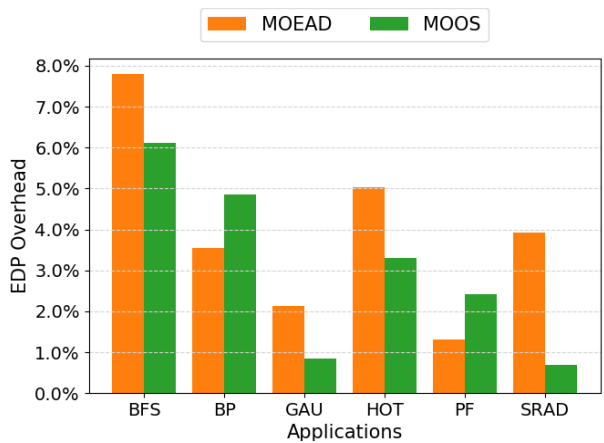

Fig. 3. EDP Overhead of MOEA/D and MOOS compared to MOELA.

From the population we acquire at the time $T_{stop}$ for the 5-obj scenario, we find the design with the lowest peak temperature for each application. We then set a temperature threshold for each application at 5% higher than this lowest peak temperature. We choose one design for each algorithm and application that has the lowest EDP within this temperature threshold. If there are no designs within this temperature threshold, we choose the lowest temperature design. In Fig. 3, we show the EDP overhead of these selected designs when setting MOELA's designs as baselines. We can see that MOELA can give better designs in terms of multiple objectives, with an EDP improvement of up to 7.7% (3% and 4% on average vs MOOS and MOEA/D, respectively). It is important to note that although this EDP improvement is modest, it comes at 8.91x speedup over MOOS and 38.83x speedup over MOEA/D for the 5-obj case. This further demonstrates MOELA's ability to find good results in a much shorter time than commonly used MOO methods.

## VI. CONCLUSION

Designing a 3D NoC-based heterogeneous manycore system is challenging as it involves searching through a huge design space and trading-off between multiple objectives. In this paper, we proposed the MOELA framework that utilizes a hybrid ML-guided local search and evolutionary algorithm approach to improve the speed and quality of the design space exploration process for emerging manycore systems. Compared with the state-of-art, the designs generated by MOELA are up to 7.7% better in EDP and up to 128x time saving in design space exploration. The proposed MOELA approach can also be utilized to more broadly improve solution quality and reduce search time for multi-objective design space searches across many other problem domains.